\documentclass[runningheads]{llncs}

\usepackage[mobile]{eccv}

\usepackage{eccvabbrv}

\usepackage{graphicx}
\usepackage{booktabs}

\usepackage[accsupp]{axessibility}  

\usepackage{hyperref}

\usepackage{orcidlink}

\usepackage{marvosym}

\begin{document}

\title{First Place Solution to the ECCV 2024 ROAD++ Challenge @ ROAD++ Spatiotemporal Agent Detection 2024}

\author{Tengfei Zhang \and
Heng Zhang \and
Ruyang Li \and
Qi Deng \and
\\
Yaqian Zhao \and
Rengang Li
}

\authorrunning{T. Zhang et al.}

\institute{Inspur (Beijing) Electronic Information Industry Co., Ltd., Beijing, P.R.China
\email{\{zhangtengfei,zhangheng10,liruyang,dengqi01,zhaoyaqian,lirg\}@ieisystem.com}
}

\maketitle

\begin{abstract}
This report presents our team's solutions for the Track 1 of the 2024 ECCV ROAD++ Challenge.
The task of Track 1 is spatiotemporal agent detection, which aims to construct an "agent tube" for road agents in consecutive video frames.
Our solutions focus on the challenges in this task, including extreme-size objects, low-light scenarios, class imbalance, and fine-grained classification. 
Firstly, the extreme-size object detection heads are introduced to improve the detection performance of large and small objects.
Secondly, we design a dual-stream detection model with a low-light enhancement stream to improve the performance of spatiotemporal agent detection in low-light scenes, and the feature fusion module to integrate features from different branches. 
Subsequently, we develop a multi-branch detection framework to mitigate the issues of class imbalance and fine-grained classification, and we design a pre-training and fine-tuning approach to optimize the above multi-branch framework.
Besides, we employ some common data augmentation techniques, and improve the loss function and upsampling operation.
We rank first in the test set of Track 1 for the ROAD++ Challenge 2024, and achieve 30.82\% average video-mAP.
\end{abstract}

\section{Introduction}
\label{sec:intro}
Accurate detection and identification of road participants, including pedestrians, vehicles, bicycles, and more, is essential for ensuring the safety of autonomous driving vehicles.
The results of such detections and identifications are pivotal for enhancing the decision-making capabilities of self-driving cars.
Consequently, the ECCV 2024 ROAD++ Challenge\footnote{https://sites.google.com/view/road-eccv2024/challenge}~\cite{singh2022road} is designed to delve into the creation of semantically meaningful representations of road scenes based on the concept of road events, with the goal of advancing autonomous driving technology. The challenge is structured around three tracks: spatiotemporal agent detection, spatiotemporal road event detection, and multi-label atomic activity recognition.

Track 1 of the challenge is focused on "Spatiotemporal Agent Detection". The objective of this track is to construct an "agent tube" for road agents, which are active objects like vehicles and pedestrians, in consecutive video frames. An agent tube is a sequence of frame-wise object detection bounding boxes. In essence, this task aligns with the concept of object tracking.

We perform a thorough analysis to identify the challenges. 
The first challenge is the detection of objects with extreme sizes. Large objects are frequently obscured, and the camera may only capture a portion of them. While small objects with few pixels are easily missed.
Consequently, these extreme-size objects increase the difficulty of detection and tracking.
The second challenge is posed by low-light scenarios. The dataset for this challenge encompasses some night scenes, where detection and tracking become arduous.
The third challenge is overfitting, which limits the model's performance on the test set. 
The fourth challenge is class imbalance. There are sufficient training samples for categories like pedestrians and cars. But the training samples are scarce for other categories, leading to suboptimal performance on rare categories. 
The last challenge is fine-grained classification. This challenge provides a more detailed categorization of vehicles, increasing the difficulty of object classification.

To enhance the detection of both large and small objects, we incorporate two detection heads for extreme-size objects.
Furthermore, we design a dual-stream detection framework that integrates a low-light enhancement stream.
This innovation utilizes the low-illumination image enhancement technique to boost the perception ability in night scenes. 
Besides, we develop a feature fusion module that harnesses the power of convolutional block attention mechanisms, thereby augmenting the model's feature representation capabilities. 
To mitigate the issues of class imbalance and fine-grained classification, we build a multi-branch detection framework.
Finally, we introduce various data augmentation techniques to mitigate model overfitting, and improve the loss function and unsampling operation.

\section{Challenges Analysis}
We identify several challenges of spatiotemporal agent detection, including:
extreme-size objects, low-light scenarios, overfitting, class imbalance, and 
fine-grained classification.

\subsection{Extreme-Size Objects}
Objects of extreme sizes are a common challenge in detection and tracking tasks.
Extreme-size objects are prone to missed detections, such as large and small objects. 

\begin{figure}[htbp]
  \centering
  \begin{subfigure}{0.29\linewidth}
    \includegraphics[width=\linewidth]{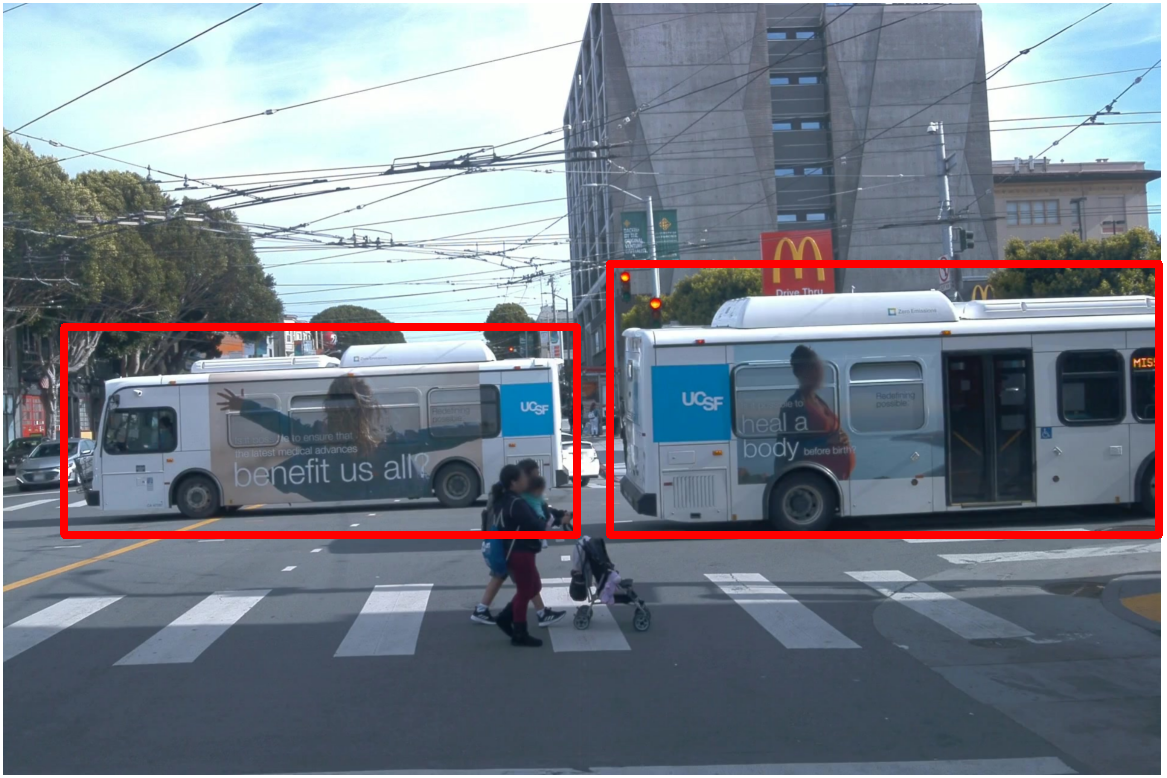}
    \caption{}
    \label{fig:extream_obj_01}
  \end{subfigure}
  \hfill
  \begin{subfigure}{0.29\linewidth}
    \includegraphics[width=\linewidth]{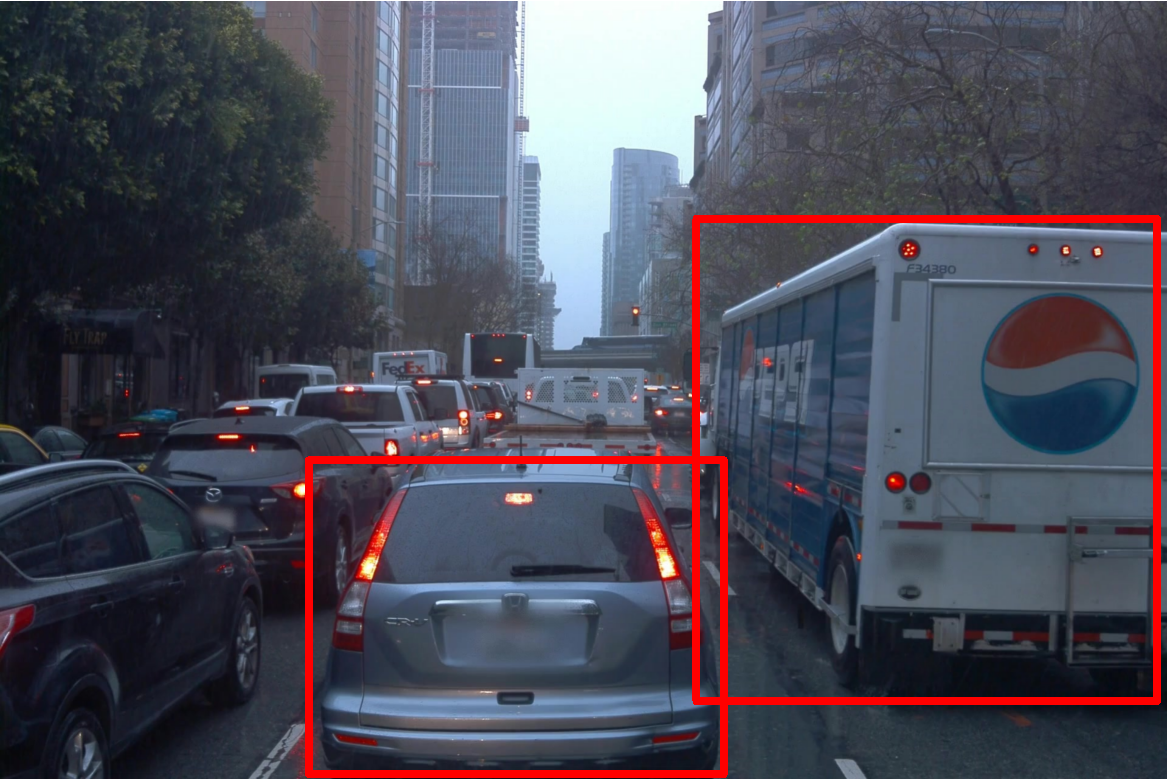}
    \caption{}
    \label{fig:extream_obj_02}
  \end{subfigure}
    \hfill
  \begin{subfigure}{0.39\linewidth}
    \includegraphics[width=\linewidth]{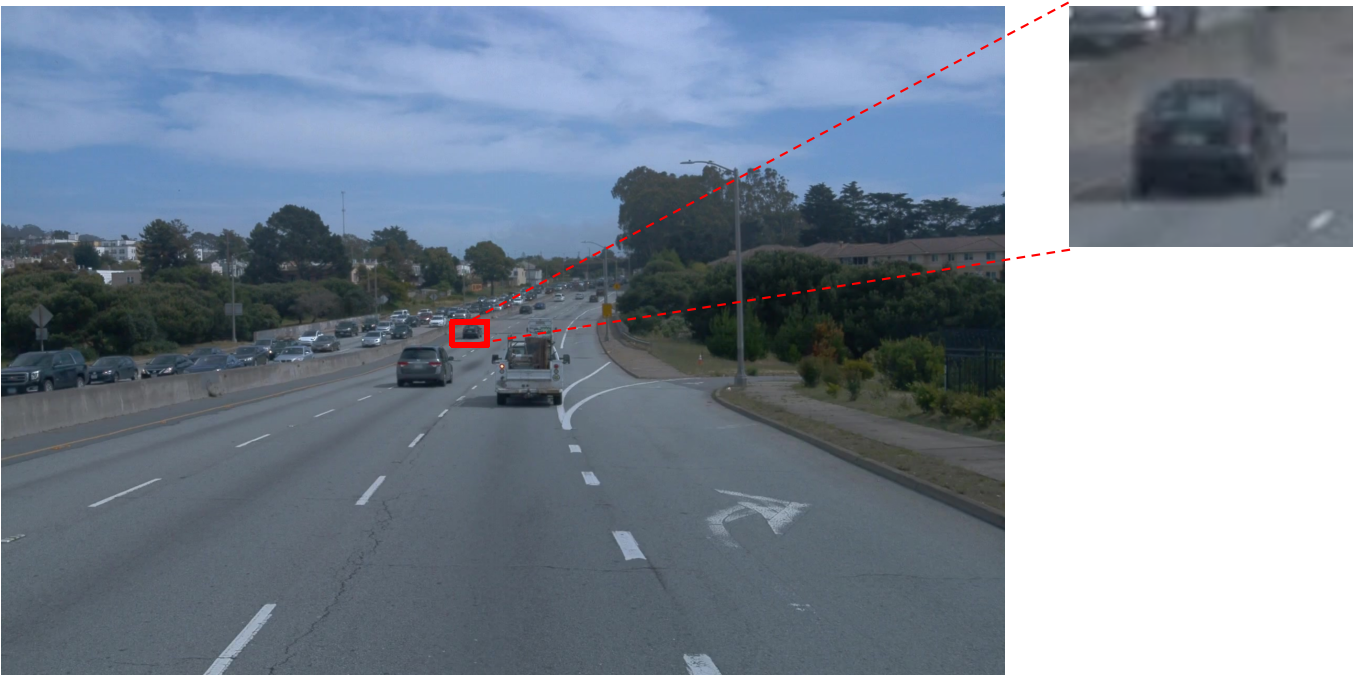}
    \caption{}
    \label{fig:extream_obj_03}
  \end{subfigure}
  \caption{Examples of extreme-size objects.}
  \label{fig:extream_obj}
\end{figure}

For large objects, the model may require a larger receptive field to capture the global information of these objects. 
As shown in~\cref{fig:extream_obj}, due to the limitations of the receptive field, the model may only capture partial information about large objects, which can result in the inability to detect these objects accurately.

Small objects are often overlooked as their image features may not stand out sufficiently, especially when the image resolution is low or the contrast between the object and its background is subtle, the features of small objects can be obscured by noise, posing a challenge for accurately detecting them.

\subsection{Low-Light Conditions}
There exist some low-light scenes, where the poor illumination greatly increases the difficulty of object detection and tracking.
Furthermore, in low-light conditions, the recognition of the fine-grained vehicle categories is more challenging.

\begin{figure}[htbp]
  \centering
  \begin{subfigure}{0.45\linewidth}
    \includegraphics[width=\linewidth]{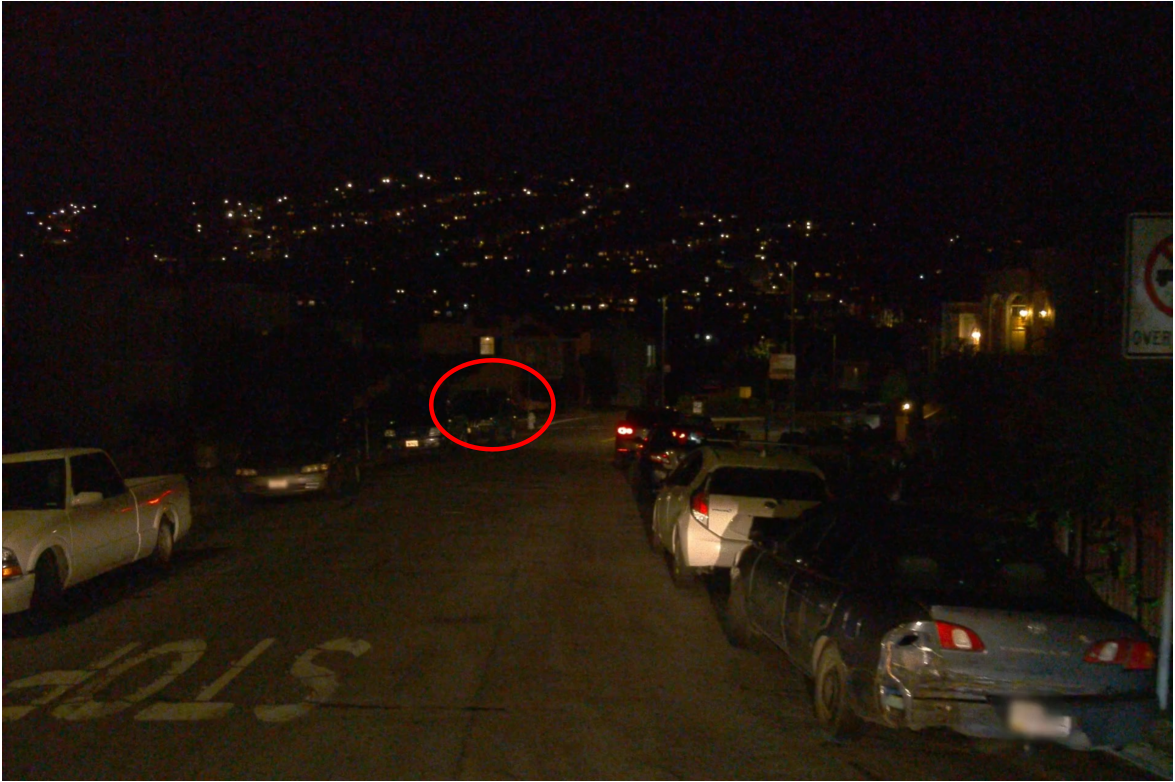}
    \caption{}
    \label{fig:low_light01}
  \end{subfigure}
  \hfill
  \begin{subfigure}{0.45\linewidth}
    \includegraphics[width=\linewidth]{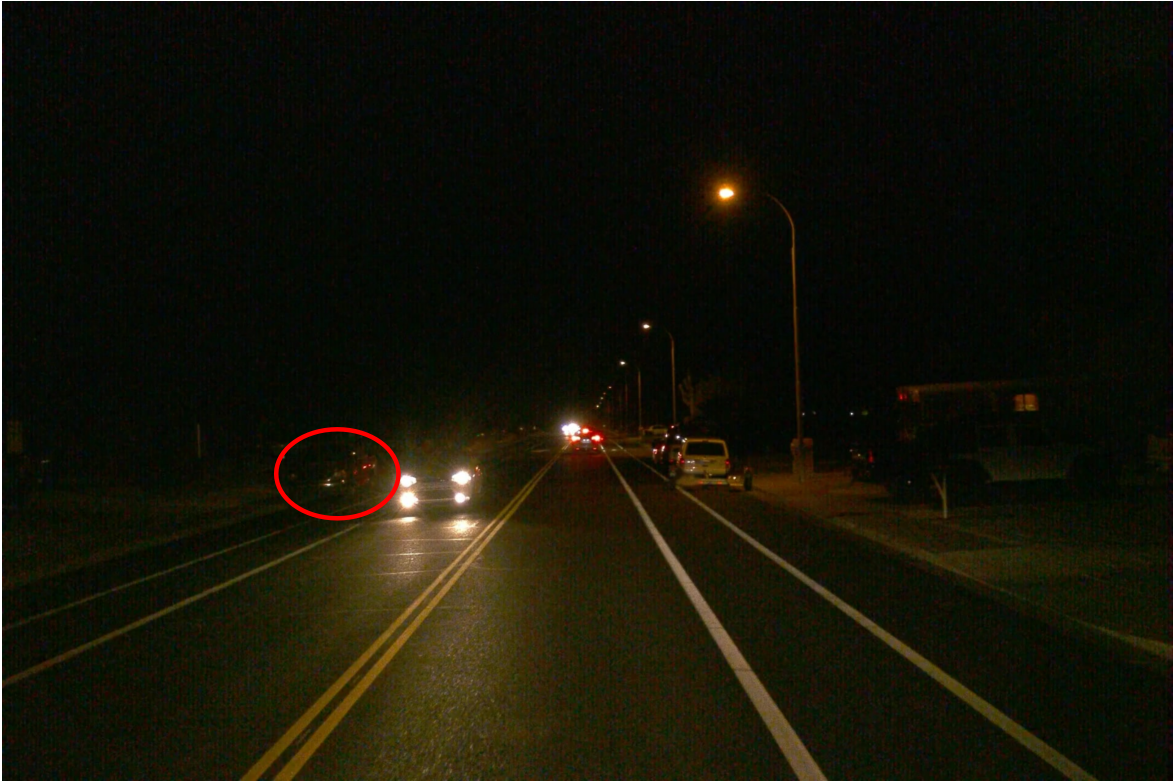}
    \caption{}
    \label{fig:low_light02}
  \end{subfigure}
  \caption{Low-light scenes.}
  \label{fig:low_light}
\end{figure}

As shown in~\cref{fig:low_light}, some vehicle objects can be identified, but differentiating between them, such as distinguishing cars from medium-sized vehicles, is challenging.
Low-light scenes constitute a certain portion of the dataset. Therefore, enhancing the model's detection capabilities in such conditions can lead to an improvement in the overall performance metrics.

\subsection{Overfitting}
Overfitting is a common issue that can stem from various factors, including limited training data and suboptimal optimization algorithms. In this task, we encounter instances of overfitting, complicating the selection of the most effective model.

\begin{figure}[htbp]
  \centering
  \includegraphics[width=0.6\linewidth]{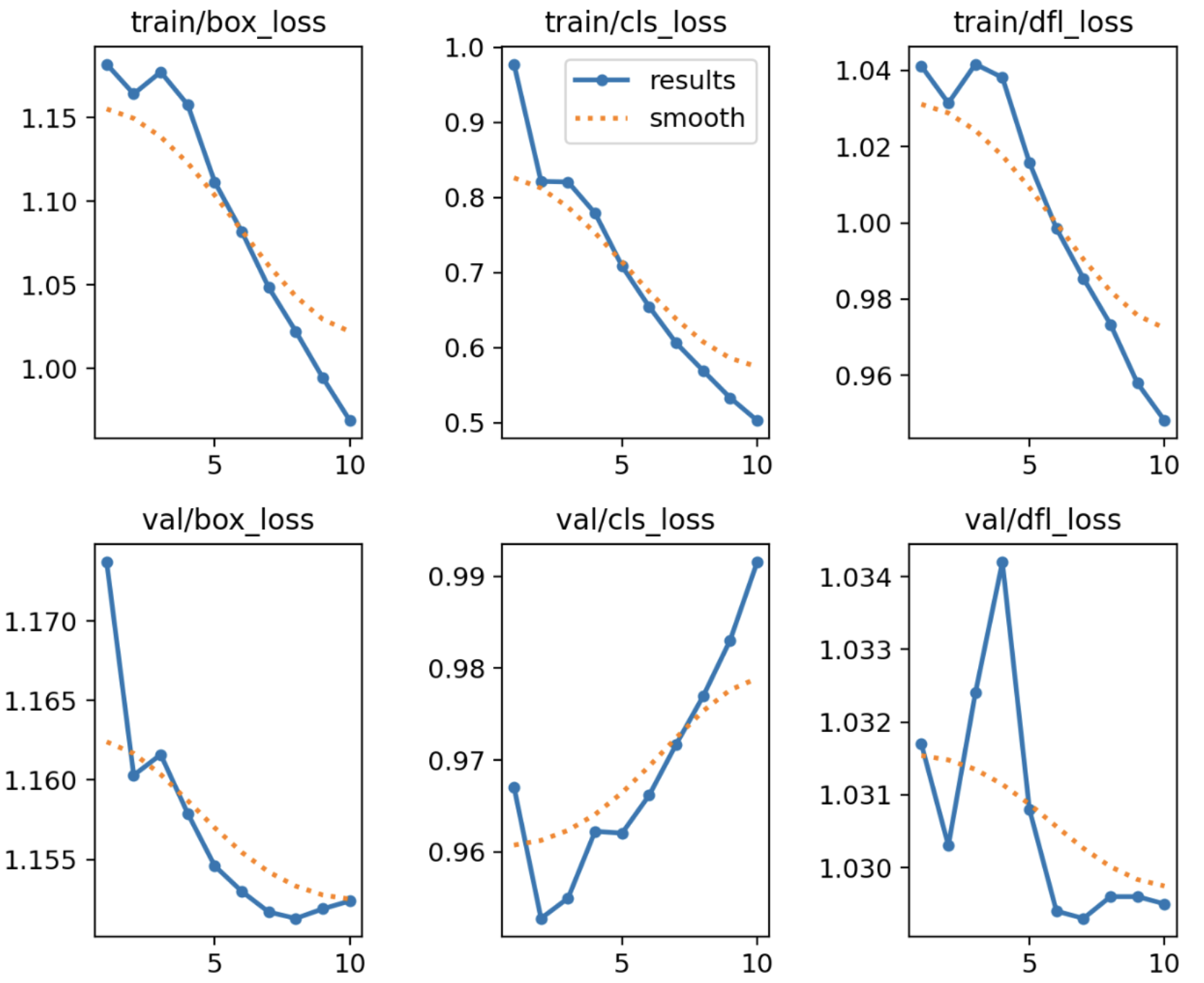}
  \caption{Loss function curves.
  }
  \label{fig:overfitting}
\end{figure}

As shown in~\cref{fig:overfitting}, the overfitting in classification loss is more pronounced than in the bounding box loss. We attribute this to the fine-grained classification for vehicle objects, which will be detailed subsequently.

\subsection{Class Imbalance}
\begin{figure}[htbp]
  \centering
  \includegraphics[width=0.4\linewidth]{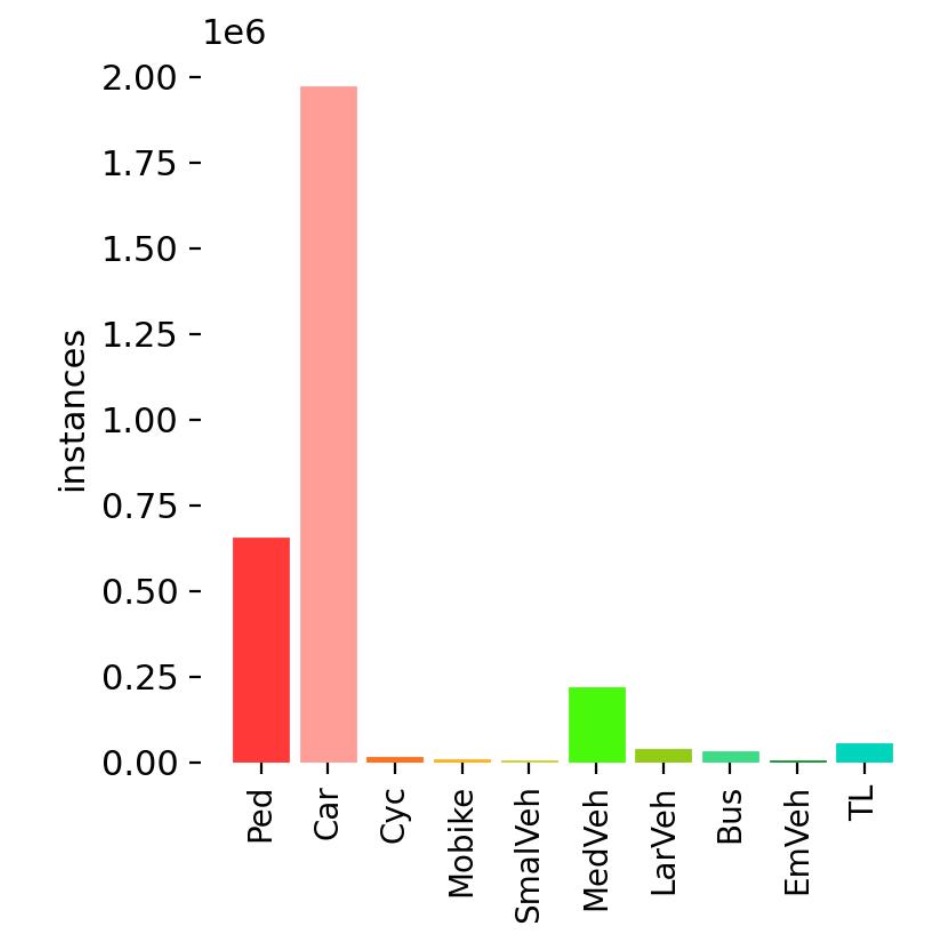}
  \caption{The number of training samples of different categories.
  }
  \label{fig:class_imbalance}
\end{figure}

As shown in~\cref{fig:class_imbalance}, there is a considerable disparity in sample counts across different object categories. 
Such data imbalance can cause models to become biased towards the majority class, thereby harming the performance of the minority class.

\subsection{Fine-Grained Classification}
\begin{figure}[htbp]
  \centering
  \begin{subfigure}{0.28\linewidth}
    \includegraphics[width=\linewidth]{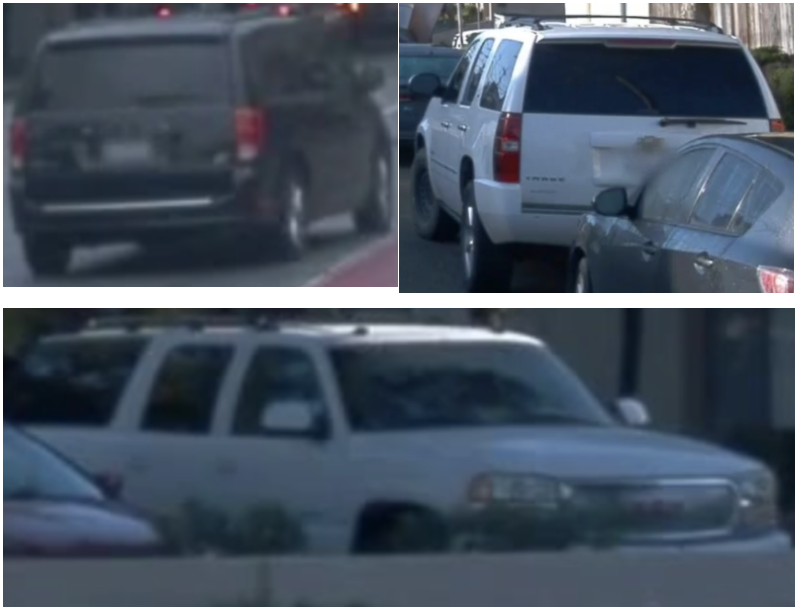}
    \caption{Car}
    \label{fig:obj_car}
  \end{subfigure}
  \hfill
  \begin{subfigure}{0.285\linewidth}
    \includegraphics[width=\linewidth]{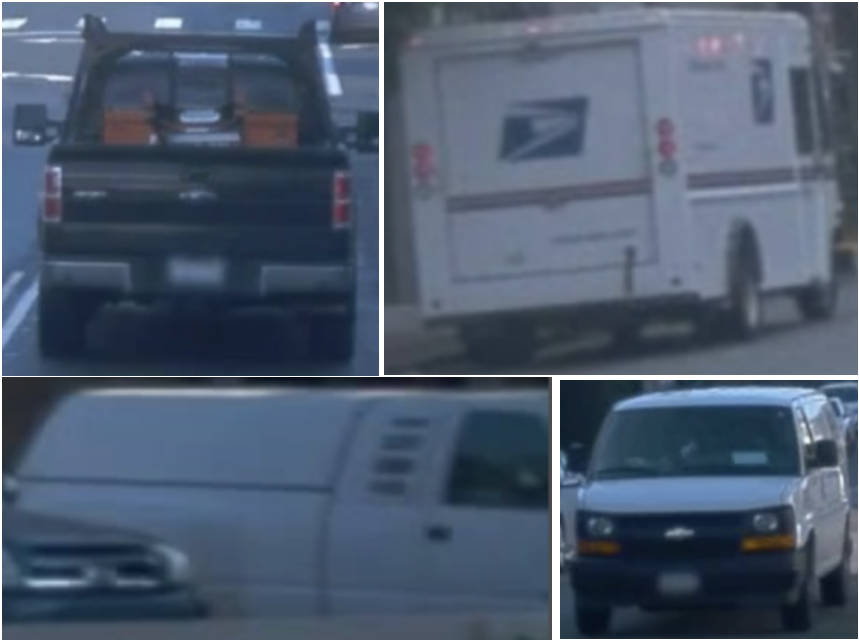}
    \caption{Medium vehicle}
    \label{fig:obj_medium_vehicle}
  \end{subfigure}
    \hfill
  \begin{subfigure}{0.19\linewidth}
    \includegraphics[width=\linewidth]{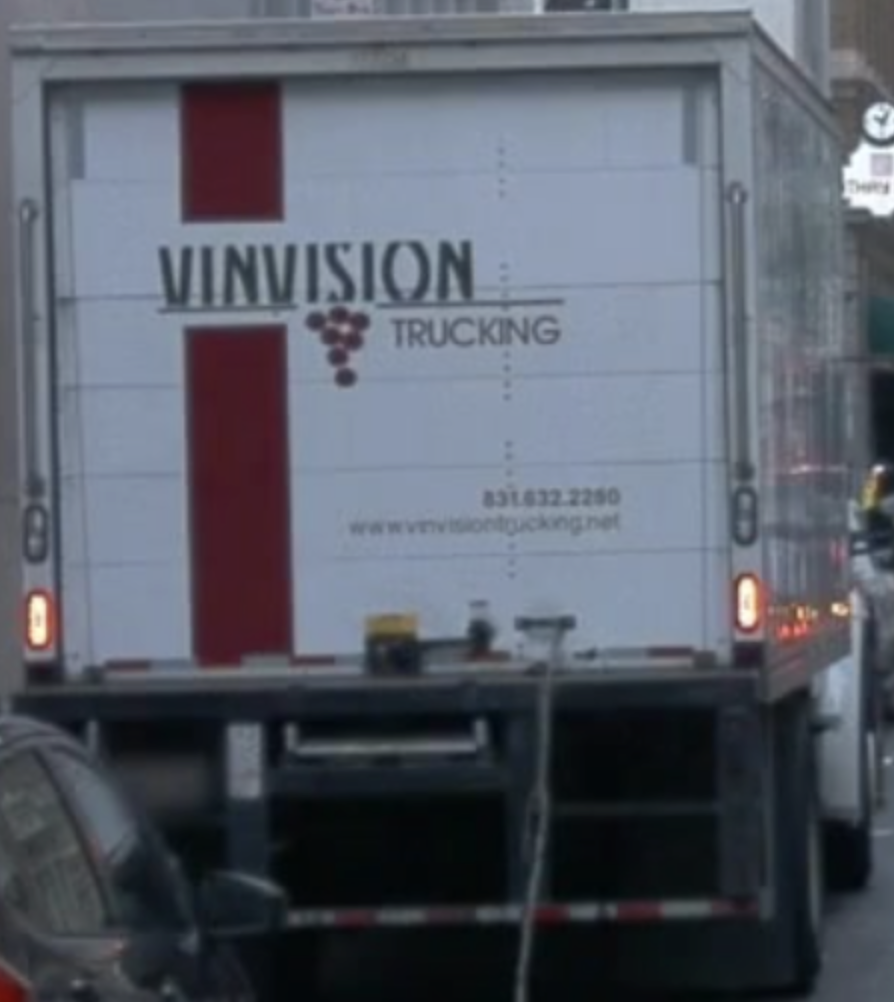}
    \caption{Large vehicle}
    \label{fig:obj_large_vehicle}
  \end{subfigure}
    \hfill
  \begin{subfigure}{0.18\linewidth}
    \includegraphics[width=\linewidth]{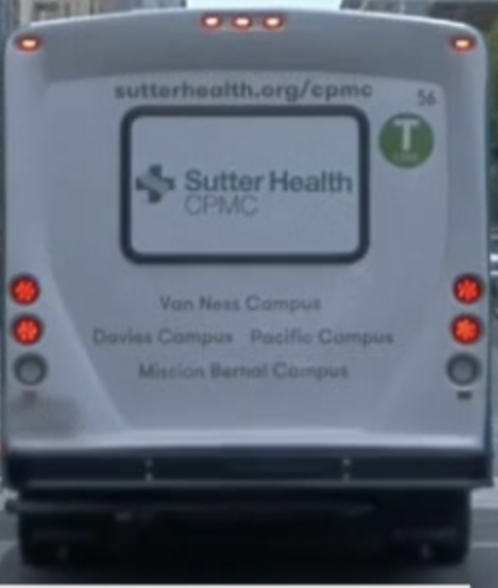}
    \caption{Bus}
    \label{fig:obj_bus}
  \end{subfigure}
  \caption{Instances of different categories.}
  \label{fig:obj_examples}
\end{figure}

The ROAD++ dataset offers a more fine-grained object categories compared to the Waymo dataset~\cite{Sun_2020_CVPR}. 
For instance, the vehicle category is further broken down into car, small vehicle, medium vehicle, large vehicle, bus, and emergency vehicle. 
Notably, we've encountered challenges in differentiating between some medium and large vehicles.
As shown in~\cref{fig:obj_examples}, 
some objects from different classes are challenging to classify, particularly when it comes to distinguishing between medium and large vehicles.

\section{Solutions}
We propose solutions to the aforementioned challenges, including extreme-size detection head, low-light enhancement, model ensembling, and pre-training and fine-tuning strategies.
We construct the model for this task based on YOLOv8~\cite{Jocher_Ultralytics_YOLO_2023}.
\subsection{Detection Heads for Extreme-Size Objects}
\begin{figure}[htbp]
  \centering
  \includegraphics[width=0.9\linewidth]{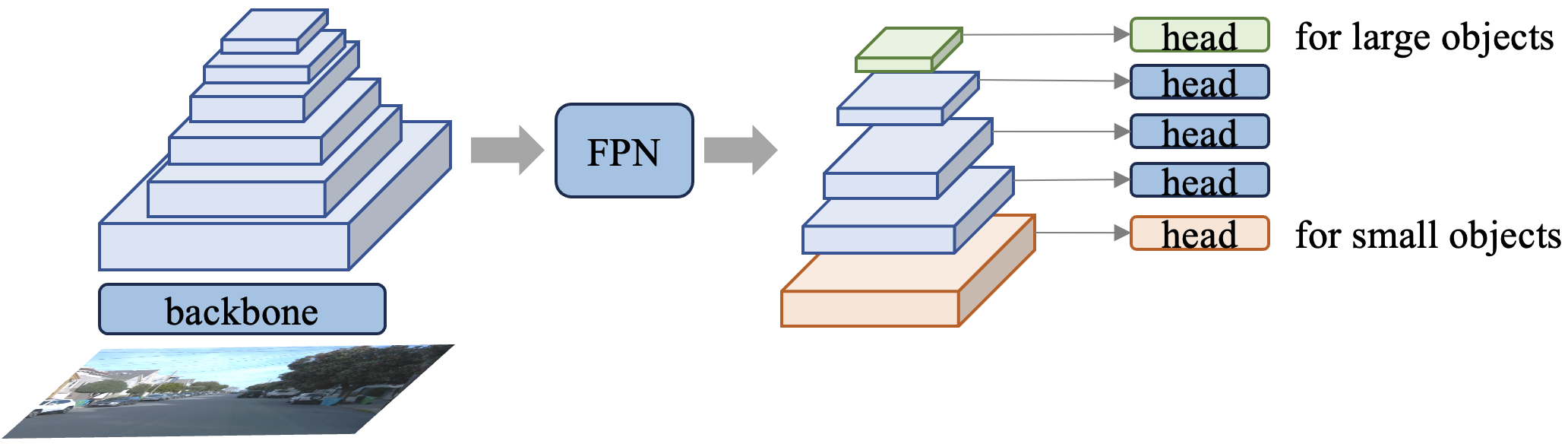}
  \caption{Extreme-size detection heads.
  }
  \label{fig:large_small_heads}
\end{figure}

For the detection of both large and small objects, we incorporate extra detection heads, as shown in~\cref{fig:large_small_heads}.

The detection head for large objects is designed using a smaller-size feature map, which provides a larger receptive field. This allows it to capture the global information of extremely large objects more effectively, enhancing the detection accuracy for such objects.

Conversely, the detection head for small objects operates on a larger-size feature map that offers higher resolution and retains more detail about small objects, thereby improving the detection results for these objects.

\subsection{Low-Light Image Enhancement}
Low-light image enhancement is a simple and straightforward method to address the challenge of low-light conditions.
There are various methods to enhance low-light images, including traditional techniques such as histogram equalization~\cite{gonzalez2009digital} and gamma correction~\cite{castleman1996digital}, as well as deep learning-based image enhancement methods~\cite{ma2022toward}. 
While deep learning approaches typically yield superior results, they often require additional training data or the extra knowledge. 
Therefore, we employ gamma correction to process the images, effectively improving the detection results in low-light scenarios.

Another challenge is the lack of information regarding which images are affected by low-light conditions. Although it is feasible to determine if an image is low-light using pre-trained models or predefined rules, we do not implement such a method. 
Instead, we opt to enhance all images indiscriminately.

\begin{figure}[htbp]
  \centering
  \begin{subfigure}{0.45\linewidth}
    \includegraphics[width=\linewidth]{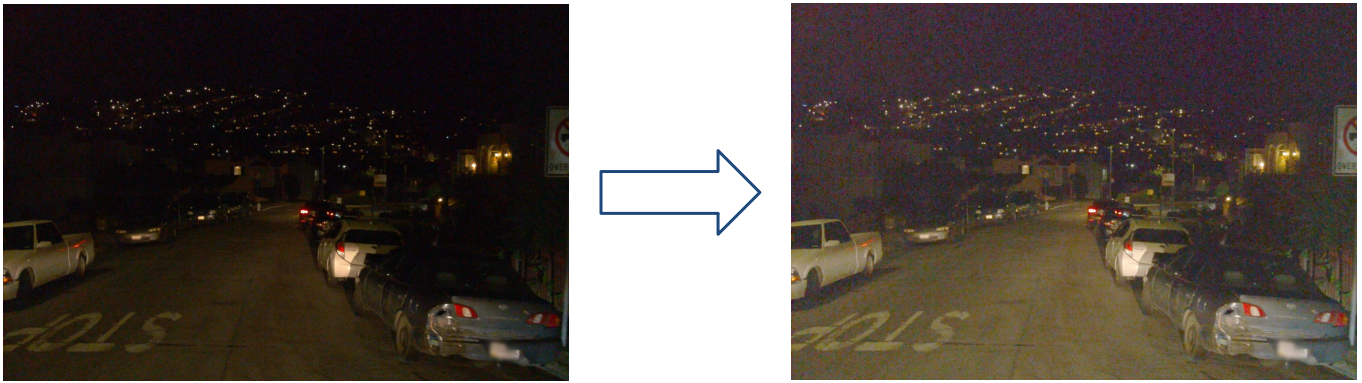}
    \caption{}
    \label{fig:night_low_light_enhance}
  \end{subfigure}
  \hfill
  \begin{subfigure}{0.45\linewidth}
    \includegraphics[width=\linewidth]{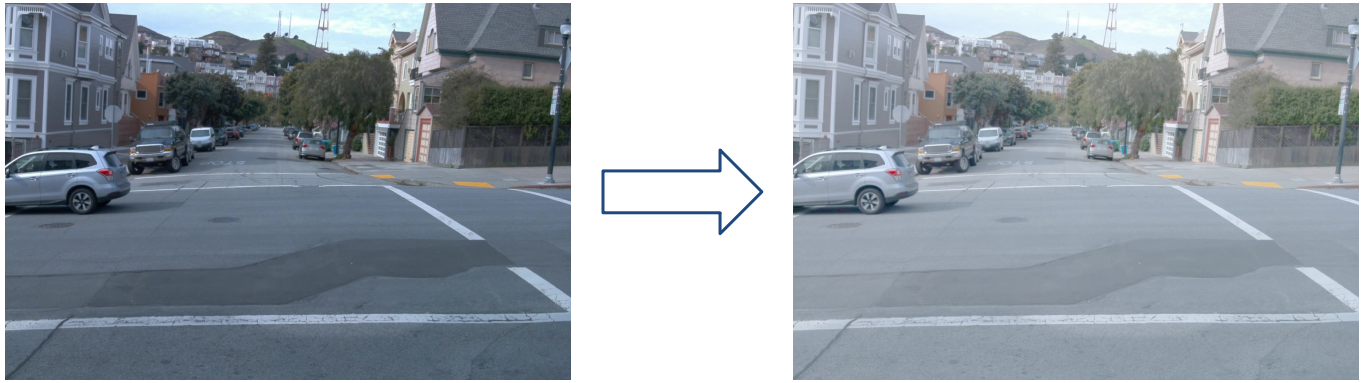}
    \caption{}
    \label{fig:day_low_light_enhance}
  \end{subfigure}
  \caption{Low-light image enhancement.}
  \label{fig:night_day_low_light_enhance}
\end{figure}

Rather than devising a specific method to identify low-light images, we choose to enhance every image in our dataset.
As illustrated in~\cref{fig:night_day_low_light_enhance}, the enhancement process is applied to both normal and low-light images alike. 

\begin{figure}[htbp]
  \centering
  \includegraphics[width=0.65\linewidth]{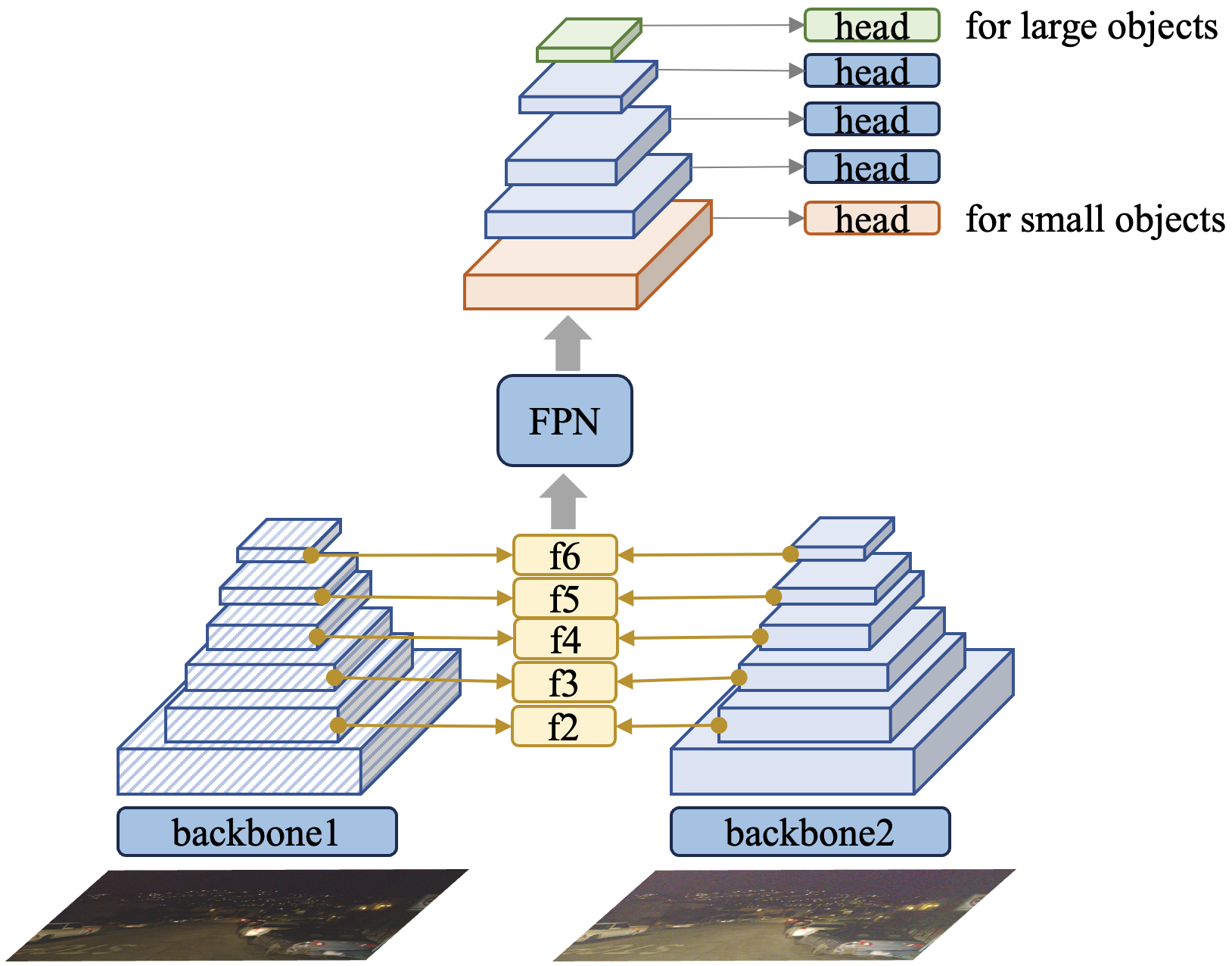}
  \caption{Dual stream detection model.
  }
  \label{fig:dual_backbone_franework}
\end{figure}

Based on the previous operations, we require distinct backbone networks to handle the original and enhanced images. 
Consequently, we develop a detection model that incorporates dual backbone networks. One of these networks is tasked with processing normal images, while the other deals with the enhanced images, as shown in~\cref{fig:dual_backbone_franework}.

\begin{figure}[htbp]
  \centering
  \includegraphics[width=0.65\linewidth]{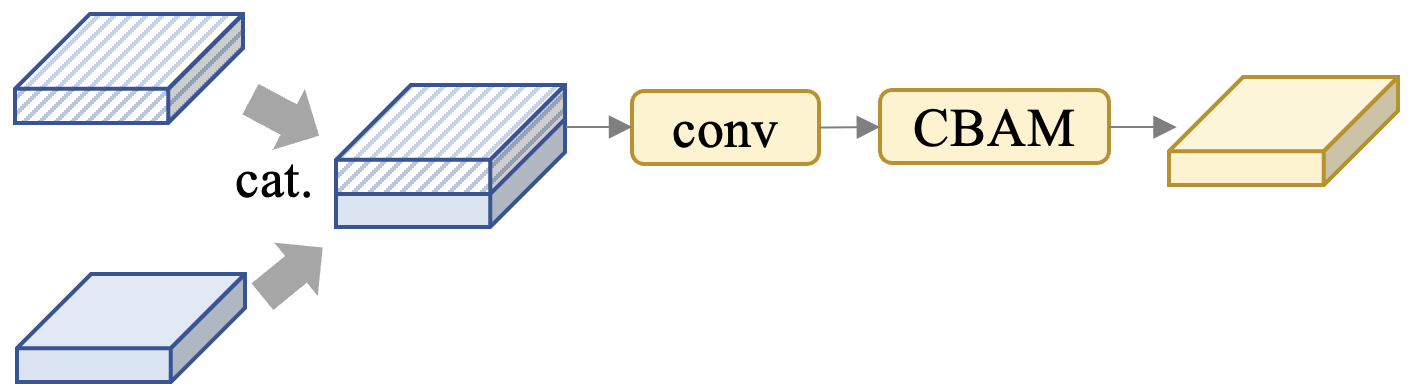}
  \caption{The feature fusion module.
  }
  \label{fig:fusion_module}
\end{figure}

Moreover, we integrate a fusion module designed to fuse features from both backbone networks, ensuring a comprehensive feature representation.
The fusion module, as shown in~\cref{fig:fusion_module}, first concatenates the features from the two backbone networks along the channel dimension. Subsequently, a convolutional layer is applied for dimension reduction, followed by a Convolutional Block Attention Module (CBAM)~\cite{woo2018cbam} attention operation.

\subsection{Model Ensembling}
\begin{figure}[tb]
  \centering
  \includegraphics[width=0.65\linewidth]{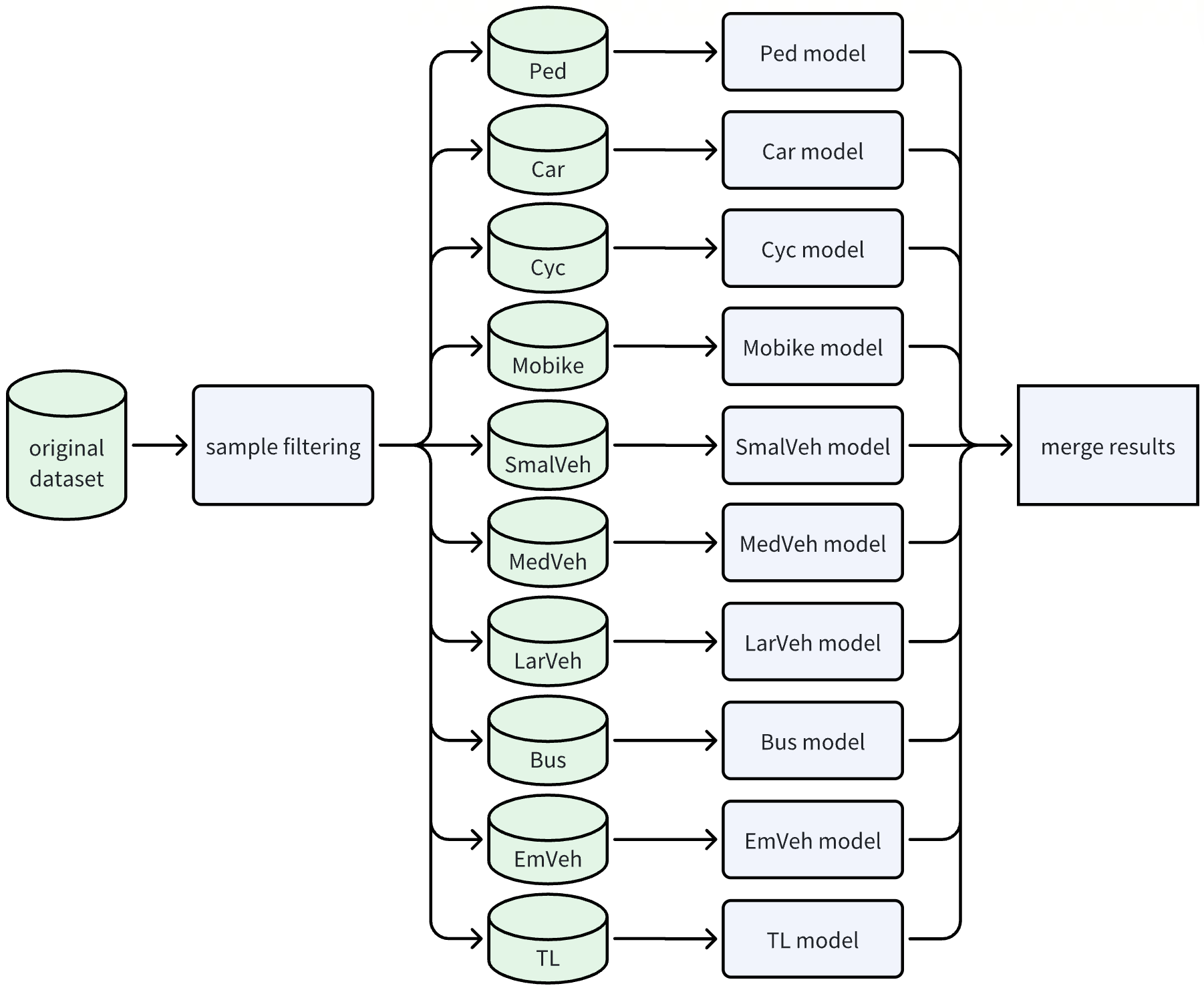}
  \caption{Model ensembling.
  }
  \label{fig:model_ensemble}
\end{figure}

As previously discussed, this task involves fine-grained vehicle classification, and is further complicated by the challenge of class imbalance.
To address these challenges, we develop a multi-branch detection framework where each individual category is assigned a dual-stream detection model, as illustrated in~\cref{fig:model_ensemble}.
By creating distinct branches for each category, our framework enhances the model's ability to learn and generalize, particularly for minority classes, leading to improved detection performance.

\subsection{Pre-training and Fine-tuning}
\begin{figure}[htbp]
  \centering
  \includegraphics[width=0.45\linewidth]{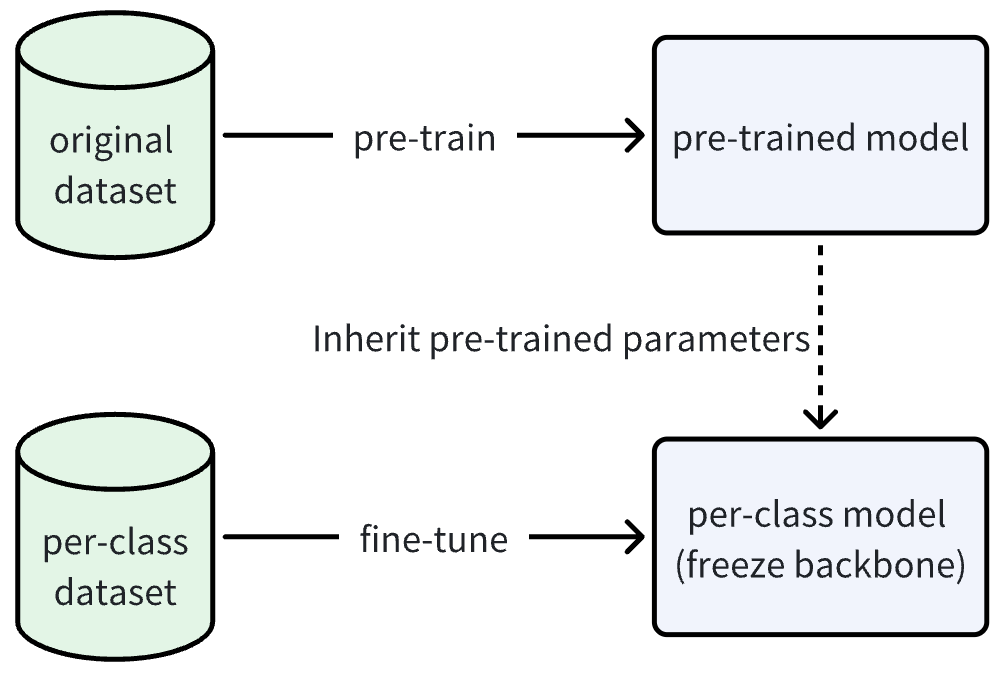}
  \caption{The pipeline of pre-training and fine-tuning.
  }
  \label{fig:pre_train_finetune}
\end{figure}

As shown in~\cref{fig:pre_train_finetune},
we utilize a "pre-training and fine-tuning" strategy to refine the aforementioned multi-branch detection framework. 
Firstly, we pre-train the model using data from all object categories.
Following this, we construct the multi-branch detection framework based on the pre-trained model and freeze the parameters of the backbone network. We then fine-tune the model on single-category data that corresponds to each branch.
Moreover, during the fine-tuning phase, we meticulously adjust the ratio of positive to negative samples to minimize false detections, 
especially for categories with limited samples, such as emergency vehicles and small vehicles.

It is worth noting that when training the models for the categories of car, medium vehicle, and large vehicle, we respectively incorporate a certain number of samples from the other two categories as negative samples. 
This is because the classification of these three categories is challenging.

\subsection{Other Solutions}
\subsubsection{Data Augmentation}
\begin{figure}[tb]
  \centering
  \begin{subfigure}{0.5\linewidth}
    \includegraphics[width=\linewidth]{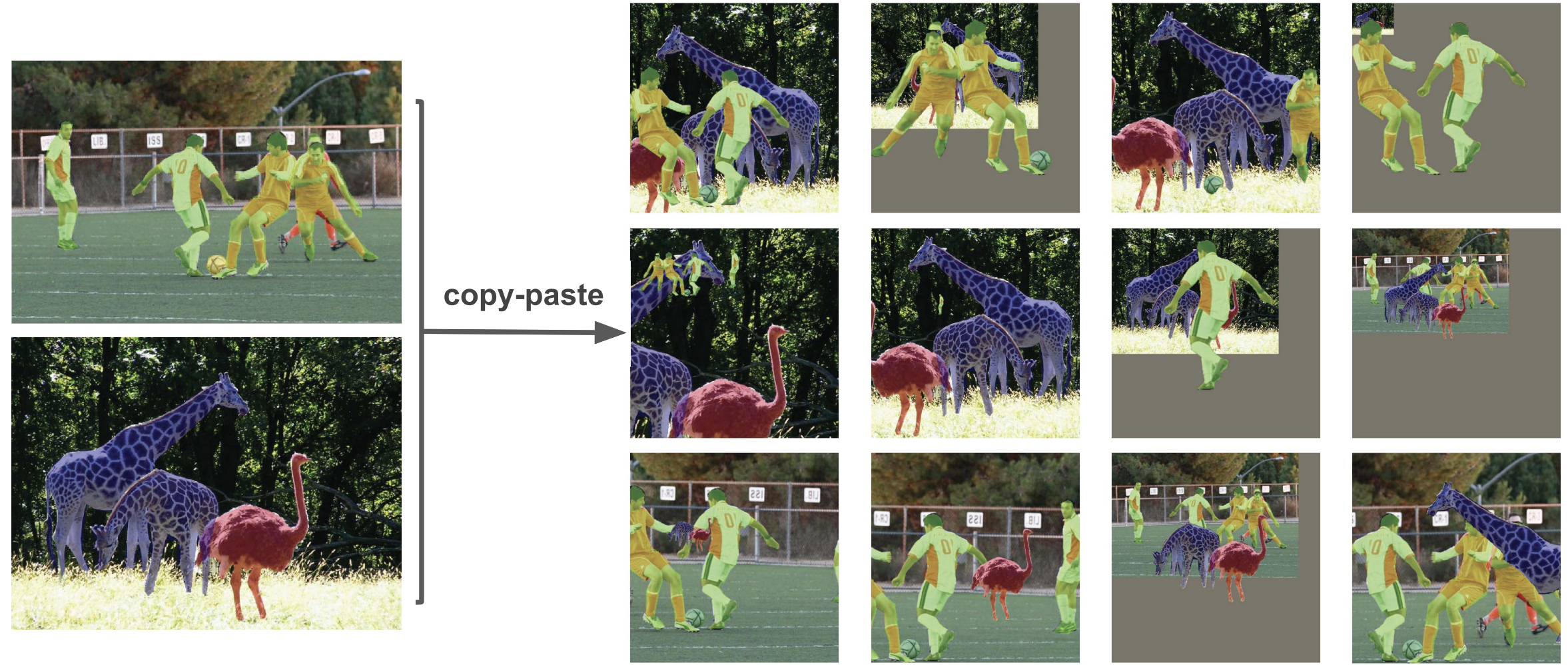}
    \caption{Copy-paste~\cite{ghiasi2021simple}}
    \label{fig:copy_paste}
  \end{subfigure}
  \hfill
  \begin{subfigure}{0.35\linewidth}
    \includegraphics[width=\linewidth]{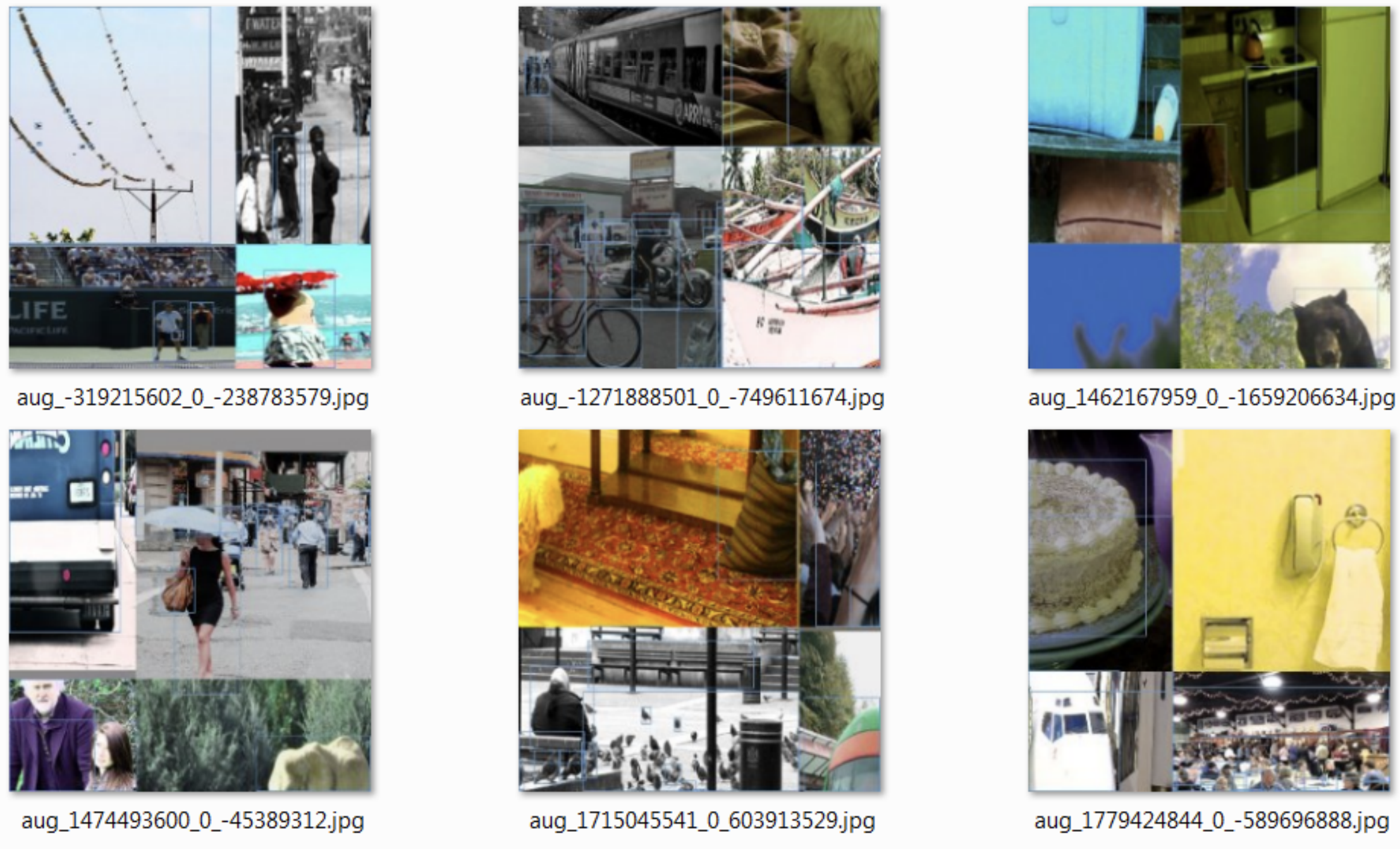}
    \caption{Mosaic~\cite{bochkovskiy2020yolov4}}
    \label{fig:mosaic}
  \end{subfigure}
    \hfill
  \caption{Data augmentation.}
  \label{fig:data_aug}
\end{figure}

To address the overfitting issue, we employ several standard data augmentation techniques, such as Copy-paste~\cite{ghiasi2021simple}, Mosaic~\cite{bochkovskiy2020yolov4}, and Mixup~\cite{zhang2017mixup}, and some examples are depicted in~\cref{fig:data_aug}.
These data augmentation operations can increase the diversity of training data, thereby mitigating the overfitting problem.

\subsubsection{Loss Function}
We use the MPDIoU (Minimum Point Distance based IoU) loss to train the model, which is a novel bounding box similarity comparison metric based on the minimum point distance, and can significantly improve object detection performance.

\subsubsection{Upsampling Operation}
We replace the original upsampling operation with the DySample~\cite{liu2023learning} upsampling operation. 
DySample is a light-weight upsampling method and can improve the upsampled features with learnable parameters.

\section{Experiments}
\label{sec:results}

\subsection{Implementation Details}
In our experiments, we utilize YOLOv8 as the base model. For the car and pedestrian categories, we construct the model based on YOLOv8x. For other object categories, we use YOLOv8m to build the model.

During the pre-training phase, the batch size is set to 32, with an initial learning rate of 0.005, using SGD as the optimizer, and train the models for 30 epochs.
In the fine-tuning stage, the parameters of the backbone network are frozen, the batch size is 32, the initial learning rate is set to 0.0005, with SGD as the optimizer, and train the models for 20 epochs. Data augmentation strategies including Copy-paste, Mosaic, and Mixup are closed in the last 5 epochs.

Besides, we split the training data into a training subset of 75\% and a validation subset of 25\%. We train the model using the training subset and use the validation subset to find the optimal number of training epochs. Finally, we train the model with the entire training data, and select the best model based on the optimal number of epochs to conduct prediction on the testing data.

\subsection{Results}
The results on the test set are shown in \cref{tab:headings}.
We achieve better performance by comprehensively applying the aforementioned methods.

\begin{table}[htb]
  \caption{Results.
  }
  \label{tab:headings}
  \centering
  \begin{tabular}{l|cccc}
    \toprule
    Method              & Agent@0.1      & Agent@0.2      & Agent@0.5      & Average        \\ \midrule
    ROAD Waymo Baseline & 8.96           & 5.71           & 1.37           & 5.35           \\
    Ours                & \textbf{39.57} & \textbf{34.48} & \textbf{18.41} & \textbf{30.82} \\ \bottomrule
    \end{tabular}
\end{table}

\subsection{Discussion}
Our solutions can be further improved.
Firstly, our solutions primarily focus on improving testing metrics rather than computational efficiency. However, this does not meet the real-time requirements of autonomous vehicles. Therefore, how to enhance computational efficiency should be explored in the future.
Additionally, the current spatiotemporal agent detection task is 2D. In the future, it can be extended to 3D space, taking into account multi-view cameras and multi-modal sensors, which should provide greater assistance for autonomous driving.

\section{Conclusion}
In response to the challenges of extreme-size objects, low-light conditions, overfitting, class imbalance, and fine-grained classification in this task, we introduce many optimization methods, including the extra detection heads for small and large objects, low-light enhancement and dual backbone model, model ensembling, optimization of pre-training and fine-tuning. Besides, we improve the detection model by employing mant data augmentation techniques, introduce advanced loss function and upsampling operations.
Our solutions significantly improve the testing results. 
Moreover, our solutions can be used independently, thus offering good scalability for solving similar issues.


%
%
\bibliographystyle{splncs04}
\bibliography{main}
\end{document}